# Decision Support Systems (DSS) in Construction Tendering Processes

Rosmayati MOHEMAD[1], Abdul Razak HAMDAN[2], Zulaiha ALI OTHMAN[2] and Noor Maizura MOHAMAD NOOR[1]

[1] Department of Computer Science, Faculty of Science and Technology, University Malaysia Terengganu
Kuala Terengganu, 21030, Terengganu, Malaysia

[2] Faculty of Information Science and Technology, Universiti Kebangsaan Malaysia
Bangi, 43600, Selangor, Malaysia

**Abstract**
The successful execution of a construction project is heavily impacted by making the right decision during tendering processes. Managing tender procedures is very complex and uncertain involving coordination of many tasks and individuals with different priorities and objectives. Bias and inconsistent decision are inevitable if the decision-making process is totally depends on intuition, subjective judgement or emotion. In making transparent decision and healthy competition tendering, there exists a need for flexible guidance tool for decision support. Aim of this paper is to give a review on current practices of Decision Support Systems (DSS) technology in construction tendering processes. Current practices of general tendering processes as applied to the most countries in different regions such as United States, Europe, Middle East and Asia are comprehensively discussed. Applications of Web-based tendering processes is also summarised in terms of its properties. Besides that, a summary of Decision Support System (DSS) components is included in the next section. Furthermore, prior researches on implementation of DSS approaches in tendering processes are discussed in details. Current issues arise from both of paper-based and Web-based tendering processes are outlined. Finally, conclusion is included at the end of this paper.

## 1. Introduction

Planning for construction project involving millions ringgit is a challenging and complex task faced by multiple parties such as clients, consultants and contractors. The successful execution of a construction project is heavily impacted by making the right decision at the right time. The construction project life cycle involves three stages namely as pre-construction, construction and post-construction [16]. Tendering processes is in the first stage [21].

Tendering processes in construction industry are fragmented and different with the other domain of tendering practices [24]. Managing tender procedures is often very complex and uncertain, involving coordination of many tasks and individuals. Different individuals have different priorities and objectives. It increases the need for efficient decisions from clients, consultants (acting on behalf of clients) or contractors. For example, contractor is required to make decision to determine either to participate in the tender or not and selection of the most qualified contractor to run the project by clients. Conventionally, decision makers tend to make decision based on a blend of their intuition, subjective judgement which based on past experience and emotion. This kind of practice does not guarantee consistent decision and lead to be bias. In complex environment, undetermined and insufficient external information as input during making decision could trigger to unfair, incomplete and poorly constructed result [30]. Meanwhile, despite the existing of diverse Web-based tools for supporting online tendering, current approaches do not include with decision support tools. The adoption of ICT in tendering processes does not help in making fair and structured decision. Thus, Decision Support Systems (DSS) play a great important role in order to ensure the confidentiality, greater transparency and healthy competition in tendering.

DSS are computer-based systems that assist business and organizational in complex decision-making environment. According to Marakas [32], DSS manage and process input of unstructured or semi-structured problems in order to support rather than replace decision-making process. The final selection in decision-making is under human authority. DSS act as supporting tools in assisting users by giving suggestions especially when fragmented information and complex problem involved.

Aim of this paper is to give a review on current practices of DSS technology in construction tendering processes. The organisation of this paper is structured as follows. Section 2 presents in general current practices of construction tendering processes. General concepts and components of DSS are outlined in Section 3. Further, Section 4 reviews on the implementation of DSS in tendering practices. Current challenges faced in DSS are discussed in Section 5. Finally, Section 6 concludes with a summary of this paper.

## 2. Tendering Processes in Construction Industry

As defined by Halaris et al. [36], tendering is the list of processes to produce, display and manage tender documents by client or consultant. It also involves action to perform bidding by interested contractors in order to win the contract by responding to tenders with their capabilities and skills formation. Meanwhile, according to Ng et al. [37], tendering processes begin with the analysis to ensure tender specification meets with end users need, followed by contractor selection, tender invitation and ending with contract awarding and contract monitoring. The period of tendering processes is defined to begin with tender preparation and to end with tender completion.

2.1 Types of Tender Method

In construction industry, there are three different types of tender including open tender, restricted tender and negotiated tender. These types of tender are largely practiced in Europe [26], Nigeria [4], Malaysia [38, 39], and Turkey [28]. Table 1 differentiates each type of tender.

Table 1: Types Tender Method

| Types of Tender | Description |
| --- | --- |
| Open Tender | All interested contractors are invited to submit tenders. This method opens the opportunity for new or unfamiliar contractor to compete for the contract [4]. |
| Restricted / Selected Tender | Only invited/selected contractors are allowed to bid the tender. This approach makes competition among contractors lesser. Normally, it is applied when construction contract needs special expertise and high technology. Those contractors who filled these requirements are invited. Most organizations in UK and many countries as well adopt this method [11]. |
| Negotiation Tender | Client consults the chosen contractors and negotiates the term of contract with them. Normally, it is applied under special circumstances such as:<br>• when the construction contract is too complex or unique to identify technical and financial properties,<br>• when there is an emergency situation that require the project to be completed in a short time,<br>• when there is only one contractor that fulfils the contract requirement in terms of expertise and technology,<br>• when involves security project of national importance |

2.2 General Tendering Processes

Tendering processes in construction sector involve set of parties. They are client and consultant as contracting authorities as well as interested organizations or contractors. Usually in many countries, the government is the largest construction client. Client is also identified as project owner [28], consultant as info broker and contractor as provider or tenderer [36]. Despite the use of different terms for describing actors in tendering, the role of each of them remains similar.

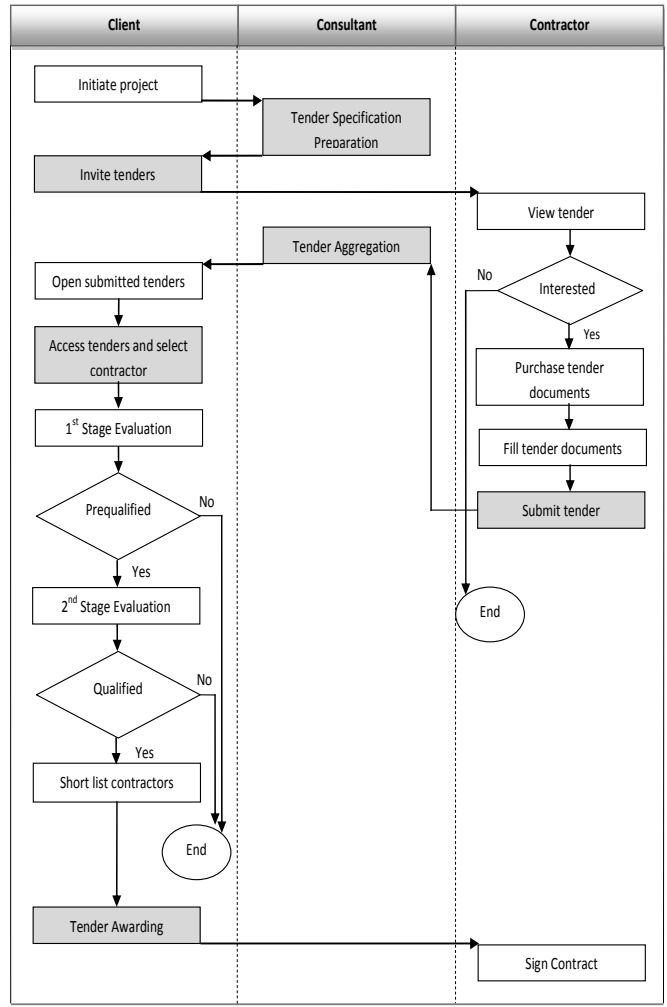

Fig. 1. General tendering processes in construction industry according to client, consultant and contractor perspectives. The grey boxes represent main phases in tendering. It starts with tender specification preparation, tender invitation, tender submission/tender bidding, tender aggregation, tender assessment and tender awarding.

Conventional tendering involves receiving, checking, copying, and distributing processes which are implemented based on paper-based. The approach remains predominant in developing countries such as Malaysia [24, 40], Nigeria [41], and Oman [42]. Strong emphasize on the need of ICT in Malaysian construction tendering were discussed by Lou and Syed Alwee [43] and Mastura et al. [44] through their analysis according to Technology Readiness Index. However, until now the real-world implementation of Web-based technology in Malaysian construction tendering faced the biggest challenge to realism. Fig. 1 summarizes the general tendering processes according client, consultant and contractor perspectives. Most countries in different region such as United States, Europe, Middle East and Asia have similar execution of tendering.

The process starts when the client initializes a construction project. Client hires consultant to prepare tender specification by conducting feasibility study. It involves typical activities such as determining estimated cost, time for project completion and procurement procedures. Consultant also responsible to prepare tender documents including qualification documents and bill of quantities. These documents require an approval from client. In order to ensure fair competition and transparency, it is compulsory for client to display notice call for tender invitation. Normally, client advertises the notice within a period of time on printed media such as newspaper, public media or website. Construction tendering procedures under European Union legislation requires a notice to be displayed in the official Journal of the European Commission [26]. Meanwhile in Nigeria, contractors are invited for tender submission through advertisement in major national publications [41]. The procedure also known as Request for Tender (RFT), Notice to Bidders or Invitation to Tender (ITT).

Subsequently, contractor who intends to bid for the tender will purchase the tender documents and give appropriate information required, and then submits the completed tender before reaching the deadline. Normally, tender documents consist of information regarding on instruction to tenderers, conditions of contract, technical specifications, drawings, bill of quantities (BQ) and list of forms to be completed by the contractor. All the submitted tenders will be aggregated and opened after the deadline.

All the tenders received will be assessed in order to select the most eligible contractor. Assessment is the most crucial stage in tendering processes because it contributes to the decision in choosing the most qualified contractor to win the tender. It is an unstructured and complex process where multi criteria need to be considered during assessment such as bid price, time for project completion, financial capability, work experience, technical staff available, equipment facilities and current list of works. These criteria contribute to identify potential contractor that capable to deliver high quality of completed service, within time and under budget allocated [28]. Usually, the assessment is carried out by the group of appointed committees. This task is under responsibility of client or his representative and tender administration committee as practices in China [45]. There are two stages of assessment before qualified contractors will be listed out.

The first stage of evaluation is the prequalified stage for contractors where they are primarily evaluated for their minimum capabilities to satisfactorily carry out the contract if they are awarded [11]. In this paper, we called this stage as prequalification phase. This stage helps to screen out ineligible or unsuitable candidates thus minimizing the amount of available contractors in the list. Different countries or organizations might evaluate dissimilar criteria. For example, in Hong Kong and Malaysia, the criteria for the first stage evaluation are based on basic requirements specified in tender notification or advertisement such as sufficiency of tender documents submitted (compulsory and supporting documents), consistency of information provided, adequate financial resources and others. It is different in other countries practices like in Turkey and Nigeria where their first evaluation are based on detailed technical evaluation. Contractors who have passed these requirements according to certain standard value will be preceded to the next stage of evaluation.

The second stage of evaluation normally comprises a detailed technical evaluation of prequalified candidates. For instance, in Hong Kong and Malaysia, contractors are evaluated according to critical criteria such as such as working experience, current work performance, technical staff, plant and equipment as well as estimated project duration. There are some evaluations practices only consider on the lowest bidding price as implemented in Nigeria and Turkey.

The qualified applicants who have fulfilled all the requirements and conditions are ranked in ascending order according to their points/credits. Contractors who do not fulfil any stage of evaluation will be automatically removed from the list. Finally, the result will be announced by client and display it either on printed media or website. Generally, the construction contract is awarded to the contractor with the highest score.

## 2.3 Web-based Tendering

The advent of Information and Communication Technology (ICT) has opened up a broad exploration to the use of Web-based technology in tendering processes. The necessity of devising strategies to automate current business processes in order to incorporate the technology in day-to-day business processes is become crucial [38, 46]. This can be seen from various existing electronic tendering applications in many countries such as United States, Canada, Australia, Singapore, Japan, Europe, and Taiwan [47]. Online tender management systems have been developed using Internet technology such as ePerolehan [48], Tender Direct [49], e-Construction [50], MERX [51], e-Procurement [52], e-Procurement System [53], Public Contracts Scotland [54], Tender Electronic Daily [55], UK Tenders Direct [56], FACNET [47, 57], and JETRO [58]. These online applications purposely designed for managing government tenders. Table 2 compares electronic tender applications that have been developed in several countries such as Malaysia, Canada, Andhra Pradesh, Chhattisgarh, Europe, United States, Australia, and Japan based on their preferences in tendering process.

Most of public sectors offer tender advertisements, online tender forms as well as download and upload related tender documents through online Web services. However, none of these applications include evaluation of tender documents electronically and with support of decision-making process. This only benefited clients and consultants for example in terms of reducing printing costs but the evaluation part requires group of decision makers to do the assessment process manually. Lack of computerized evaluation tools in current Web-based tendering applications requires decision maker to manually screening for criteria to be evaluated for each tender documents. It is impractical and time consuming for human to manually process the information. Thus, the adoption of ICT in construction industry remains low and at the same time encourage to the increasing use of large volume of unstructured tender documents.

## 3. Decision Support Systems (DSS)

The main purpose of developing DSS is to support decision makers in decision-making process in order to handle complex problem environment. DSS have been applied in various research area of applications as discussed by Power [59]. This section explains the overview of DSS technology in terms of its definition and categories.

### 3.1 Definition of DSS

DSS definition is evolving from theory to practice as well as the improvement of various supporting technologies such as minicomputers and user friendly software applications. There is no standard and universal definition of DSS as different people with different background have different views on DSS [60].

Different researchers used a variety of overlapping terms such as online analytical processing (OLAP), executive information system (EIS), group DSS, knowledge discovery systems and business intelligence (BI) for computer-based system which capable to support decision maker or manager in making decision instead of DSS. Less use of DSS label in trade journals and vendors website show the emergence of other new marketable terms [61]. Thus, DSS often passed through some stages of development with name changes. Kopackova and Skrobackova [62] argued that scientific researchers frequently used DSS term whilst vendors defined BI term and both terms refer to any tools with decision-making functionality.

Table 2: Summary of Electronic Tender Applications between countries

| Country | Electronic Tender Application | Tender Display | Online Tender Forms | Download Tender Documents | Upload Tender Documents | Online Bidding | Tender Evaluation Tools | Tender Award Notify |
|---|---|---|---|---|---|---|---|---|
| Malaysia | ePerolehan | √ | X | √ | √ | √ | X | √ |
| | Tender Direct | √ | X | X | X | X | X | X |
| | e-Construction | √ | X | X | X | X | X | X |
| Canada | MERX | √ | X | √ | X | √ | X | √ |
| Andhra Pradesh | Tender Management System | √ | √ | √ | √ | √ | X | √ |
| Chhattisgarh | e-Procurement System | √ | √ | √ | √ | √ | X | √ |
| Europe | Public Contract Scotland | √ | √ | √ | √ | √ | X | √ |
| | Tender Electronic Daily (TED) | √ | X | X | X | X | X | √ |
| | UK Tenders Direct | √ | X | X | X | X | X | X |
| United States | FACNET | √ | X | √ | √ | √ | X | √ |
| Japan | JETRO | √ | X | X | X | X | X | √ |

The term DSS initially coined by Gorry and Scott-Morton [63] in Sloan Management Review article. They claimed DSS focus to support semi-structured and unstructured decisions. Since then, growing amount of studies in terms of interpreting DSS have been discussed and argued. Donovan [64] extended DSS as the systems with the capability to deal with complex problems by providing information and necessary analysis. Another definition of DSS is defined as "computer-based to support decision making rather than to increase transaction processing and record keeping" [65]. Despite of different variation of early definitions have been proposed, there is an acceptable understanding that DSS are system supported by computer technology in which capable to provide decision analysis for ill-structured problem.

Keen [66] argued that previous definition of DSS is for specific application. He redefined the term DSS as a completed system comes out through dynamic interactions between users, designer and systems, analysis models as well as support for technology availability. Turban and Watkins [67] classified DSS as an interactive computer-based system which employ decision rules, models and database. Further, Holsapple [68] viewed DSS as a computerized system with knowledge representation and knowledge processing in order to strengthen decision-making to be more productive, agile, innovative and reputable. More recent definition of DSS is regarded to DSS with intelligent behaviour.

### 3.2 Categories of DSS

Basic DSS design consists of user interface, data management and model-based management. According to Shim et al. [69], user interface is useful to support direct communication between decision makers and the system. Friendly user interface is paramount with respect to achieve intensive interaction between decision makers and the computer. Meanwhile, data management system includes a database that stores relevant data for the situation and normally is managed by database software [1]. It also functions to store and access internal and external data. Model management supports the system with analytical capabilities by formulating data. Fig. 2 shows a schematic view of DSS as defined by Turban et al. [1]. There are five key approaches of DSS defined based on the input they can handle and type of decision processes they can support including communication-driven, data-driven, document-driven, model-driven and knowledge-driven DSS [59, 70, 71]. DSS have been applied in diverse research area of applications. In real-world practices, DSS applications could be implemented by overlapping approaches. This includes a great number of studies that have implemented DSS in construction tendering processes. Tendering is a complex problem with conflicting interests between construction managers, multiple sets of objectives, criteria and solution possibilities.

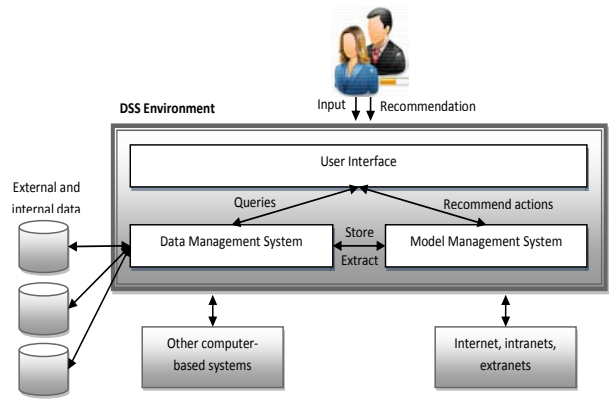

Fig. 2. A Schematic View of DSS as defined in [1].

Multiple decision makers in geographically dispersed locations can have interactive communication and collaboration via networking technologies by using communication-driven DSS approach. It is also known as group decision support systems (GDSS). Data and decision models are shared by a set of group decision makers. In some particular problems, it could be the integration of communication-driven and model-driven approach. Further, data-driven DSS allows accessing, manipulation, process and analysing large volumes of structured data using computerised techniques such as for examples, data warehousing and OLAP. It highlights to process and manipulate time-series of historical data, real-time operational data and external data [72]. Document-driven DSS is relatively new approach in DSS. It helps decision makers to access, retrieve, manage and analyse unstructured documents by integrating document databases and processing technologies.

Main function of model-driven DSS is based on quantitative approach. It is used to provide decision support with analytical model by using tools of decision analysis, algebraic, financial, simulation, optimization, statistics, stochastic and logic modelling. According to Shim et al. [69], this approach can be classified into three phases namely as formulation, solution and analysis. Knowledge-driven DSS uses knowledge inference engines to understand and solve problems domain as well as suggest actions to decision makers. Knowledge about a particular domain is stored in a reusable form with the help of different AI methods such as rule-based, statistic-based, heuristic, object-based, logic-based or induction based [59, 73].

## 4. Related Works of DSS in Tendering Processes

The potential of DSS as computerized tools in assisting decision-making process has been attracting many researchers for the last fifty years to improve decisions in tendering processes. Most phases in tendering processes involve crucial decisions that need to be made either by client, consultant or contractors. Prior researches that implemented decision-making concept in the area of construction tendering process are reviewed. Tender specification preparation, tender submission/tender bidding, tender assessment and contract monitoring are classified into communication-driven, data-driven, document-driven, model-driven and knowledge-driven.

4.1 Tender Specification Preparation

Tender specification preparation is the feasibility study to analyse and consider risk and opportunity when proposing a construction project. During preparing tender documents, several set descriptors of criteria and relative weights are identified for the evaluation purpose. Table 3 summaries several researches that have included DSS technology during preparing tender specification. Costa et al. [5] proposed an approach to structure screening, evaluation and weight using decision conferencing process. Further, Noor et al. [8] applied process modelling to support real time communication between clients and consultants during tender preparation stage. Meanwhile in model-driven approach, several analytical model have been proposed to determine budget estimation and risk analysis model including cost and risk analysis model [13], predictive modelling and Monte Carlo simulation using Crystal Ball software [19] and Analytic Hierarchy Process (AHP) [22].

4.2 Tender Bidding/Tender Submission

Bidding is the most common method in competition to win the construction tender. It is concerned with two crucial decisions. The first is decision-making strategies in project selection whether or not to bid for the job when the invitation has been received and the latter is prediction of reasonable bid price if contractors opt to bid [74]. Reasonable bid price or mark-up price is defined as the minimum price that possible to win the tender and could maximize profit at the same time [7]. Fuzziness in information on current new construction project, the client preferences, the potential competitors and the overall construction market make it a very complex process for contractor to make decision. Table 4 reviews on several researches that proposed approaches to assist contractor in making decision. In model-driven approach, Delphi method has been used to identify relevant factors that effectively contribute to decision for contractors either to take part in tender bidding or not, and estimate price to bid [2]. Bid reasoning model operates on several crucial factors from perspectives of reasoning goals that contribute to the overall decision whether or not to bid [7]. Meanwhile, fuzzy linguistic approach has been proposed by Lin and Ying-Te [12] to quantify imprecise and vague factors to make decision and recommend decision in linguistic terms. Further, case-based reasoning model was developed to estimate the risk, opportunity and competition ratings which in turn converted to estimated bid amount value using linear utility function [17].

Table 3: Prior Researches of DSS in Tender Specification Preparation

| Tendering Processes | Communication-Driven | Data-Driven | Document-Driven | Model-Driven | Knowledge-Driven |
|---|---|---|---|---|---|
| Tender Specification Preparation | Decision conferencing [5] Process modelling for online communication system [8] | NIL | NIL | Schedule and cost risk analysis model [13] Predictive modelling and Monte Carlo simulation [19] Analytic Hierarchy Process [22] | NIL |

Table 4: Prior Researches of DSS in Tender Bidding/Tender Submission

| Tendering Processes | Communication-Driven | Data-Driven | Document-Driven | Model-Driven | Knowledge-Driven |
|---|---|---|---|---|---|
| Tender Bidding/Tender Submission | NIL | NIL | NIL | Delphi [2] Bid Reasoning Model [7] | Fuzzy Linguistic Approach [12] Case-Based Reasoning and Linear Utility Function [17] |

### 4.3 Tender Assessment

Tender assessment is the most crucial stage in tendering processes and has gained extensive study from researchers in terms of the implementation of DSS technology as depicted in Table 5. It can be divided into two sub phase namely screening and evaluation. Screening is the process to ensure each candidate complies with minimal threshold before proceeding to comparative evaluation. Noor Maizura et al. [6] implemented Web-based technology with Javascript approach to control sufficiency of supporting documents attached.

Meanwhile, evaluation phase for communication-driven approach, web-based approach facilitated information transfer and collaborative environment between contractors in evaluating sub contractors [10]. Several researches in data-driven approach also support the assessment stage where databases have been developed to store large volume of data for evaluation purposes [15]. In knowledge-driven approach, several artificial intelligent techniques have been integrated into DSS models such as case-based reasoning [31], genetic algorithm and neural networks [33], binary goal programming [34] and support vector machine [35] to predict contractor performance. All these mechanisms train their analysis based on historical evaluation datasets.

Further, extensive researches have been done in model-driven approach. As early as 1995, a prototype of DSS was proposed for contractor prequalification where they used performance assessment scoring system to evaluate performance, management capability, reputation, resources, progress, competitiveness and activeness separately [20]. Sirajuddin and Al-Bulaihed [23] transformed criteria measures provided by each contractor in tabular format in order to ease evaluation procedure. Multi criteria utility theory model has been proposed to evaluate diverse set of criteria for contractor selection [25, 26]. Multi criteria decision model using AHP was applied for contractor selection where several criteria and sub criteria were equally analyse hierarchically [27]. It was also proposed for contractor selection in Turkey where contractors were required to pass two stages of evaluation, bid price and prequalification threshold filter [28]. In addition, similar model also was proposed by Al-dughaither [29] to tackle contractor selection problem in Saudi Arabia. The model is expected to reduce the possibility of inconsistency with data entry to the AHP matrices.

### 4.4 Contract Monitoring

Contract monitoring is to regularly checks on the contractor performance and progress of running project after awarding the contract. Several researches have been delivered in communication-driven, data-driven and knowledge-driven approach accordingly as shown in Table 6. Deng et al. [3] used Internet technology as the communication platform and developed to support project monitoring activity. The modules including data sharing using File Transfer Protocol (FTP) and Telnet, information exchange by email, Internet chat for real-time communication, live video-cam, and search engine for data collection purpose. Further, database technology is applied to store, query, display, and analyse data on claims for additional payments and time for completion of ongoing project [9]. In similar approach, Cheung et al. [14] made use of Web-based and database to manipulate data on current ongoing project by plotting graph to compare and measure performances. Meanwhile, data mining approach is integrated with DSS model to control and analyse cost of a project in order to determine project performances, predict trends and support in decision-making [18].

Table 5: Prior Researches of DSS in Tender Assessment

| Tendering Processes | | Communication-Driven | Data-Driven | Document-Driven | Model-Driven | Knowledge-Driven |
|---|---|---|---|---|---|---|
| Tender Assessment | Screening | NIL | Web-based Prequalification [6] | NIL | NIL | NIL |
| | Evaluation | Web-based sub contractor Evaluation System [10] | Database Development for Tender Evaluation [15] | NIL | Performance Assessment Scoring System [20] Tabulated procedure [23] Multi Criteria Utility Theory [25, 26] Analytic Hierarchy Process [27-29] | Case-based reasoning [31] Hybrid of Genetic Algorithm and Neural Network [33] Binary Goal Programming Model [34] Support vector machine [35] |

Table 6: Prior Researches of DSS in Contract Monitoring

| Tendering Processes | Communication-Driven | Data-Driven | Document-Driven | Model-Driven | Knowledge-Driven |
|---|---|---|---|---|---|
| Contract Monitoring | Total information transfer system [3] | Relational database for monitoring claims and time [9] Project Performance Management System [14] | | | Project Cost Control System Based on Data Mining [18] |

## 5. Current Issues

The exponential growth of unstructured information lead to the necessity of devising strategies to improve and enhance individual and organizational decision-making by associating with automated tool in decision systems [75]. It is one of the current research challenges in DSS since the standard tools designed for structured data analysis. Traditional DSS lack of capability to confront with dynamics and ill-defined data. Current existing decision support tools in construction tendering processes are focused on quantitative data processing where the systems are specifically analyses factual values. Quantitative system could not directly scrutinize exact problem structure from text. According to Froelich & Ananyan [76], current challenges in decision-making requires comprehensive analysis of large volumes of both structured and unstructured data.

Issuing free text documents either in paper or electronic forms (i.e contracts, policies, catalogues, certificates, transcripts, financial statements, surveys, medical reports) are inevitable in daily business for auditing, authentication, information exchange and recording purposes [38, 77-79]. A survey conducted by Building Cost Information Services (BCIS) on [80] showed that despite of active electronic tendering practices, document in hard copy is still required at the end of the process. It is because users prefer final documentation in black and white copy. Furthermore, there are potential limitations in web-based implementation as it is designed to solve specific application in which human effort is needed to maintain and modify the changes either interface or process with sufficient programming skills. The dominant use of paper-based approach is also due to the fact that Web-based applications do not meet the technological demands of end users to improve the whole processes [77, 81].

Supporting documents or attachments or appendices are essential in order to verify and acknowledge the information provided in the application. For example in Web-based tendering processes, client side form validation with Javascript approach is proposed to validate the sufficiency of the supporting documents attached [38].

Despite of the existing Web-based application, current approach requires users to manually go through the documents, check and evaluate the consistency of information in these documents with the information provided either in the compulsory documents or database. Current web-based tendering processes faces limitation in verifying tender documents in which tender documents need to be reviewed through computer monitors or in paper-based format [43, 47]. Hence, a framework is necessary to automatically detect outlier specification in terms of syntactic and semantic contents of supporting documents with respect to contextual application.

## 6. Conclusion

Traditional tendering processes in construction industry are complex and fragmented. This paper summarizes current practices of general tendering processes around the world such as Unites States, Europe, Middle East and Asia. Related researches on DSS in tendering processes have been successfully reviews in this paper. The benefit of decision support has attracted various researchers to implement the system in tendering. Despite efforts to integrate and transform the whole construction tendering processes into electronic or digital form by using Web-based Technology, the use of unstructured documents either in hard copy or digital could not be neglected in daily business processes especially for recording, authentication and information exchange purposes. Large volumes of unstructured tender documents need to be analysed. It is the biggest challenge to automate the analysis using computerised tools. It becomes harder when to automatically convert unstructured data to structured format data for input in decision-making processes.

The need to extract and represent information in machine-readable formats becomes obvious. In order to automate the tendering processes, integrating ontology in DSS model seems to be a promising approach. We proposed a framework of ontological-based extraction for decision support system in order to improve tender assessment process. For electronic government, as far as we are concern, none of the research has been found to use ontological modelling as part of decision support components. For the future research, we intend to construct domain ontology for tendering processes using standard methodology for ontology building. The ontology will represent rule-based for supporting decision-making process.

**Rosmayati M.** received her Bachelor Science degree in Computer Science from University of Technology Malaysia, Malaysia, in 2003. She obtained her Master's degree in Computer Science in 2004. Currently she is pursuing PhD study at University Kebangsaan Malaysia, Malaysia in Decision Support Systems. Her research interests are in Decision Support Systems and Ontology. Now, she is attached with University Malaysia Terengganu, Malaysia as lecturer.

**Abdul Razak H.** is a professor in System Management and Science, Faculty of Information Science and Technology, University Kebangsaan Malaysia, Malaysia. Currently he is the dean of this faculty. He received his BSc degree from University Kebangsaan Malaysia in 1975. He obtained his Master's degree from University of Newcastle Upon Tyne, United Kingdom in 1977 and PhD in Artificial Intelligence from Loughborough University of Technology, United Kingdom in 1987. His research interests include ICT & Strategic Policy, Intelligent Decision Support, Medical Data Mining and Jawi Pattern Recognition.

**Zulaiha A. O.** is an Associate Professor in Faculty of Information Science and Technology, University Kebangsaan Malaysia. She received her BSc degree from University Kebangsaan Malaysia, Malaysia. She obtained her Master's and PhD from Univrsity of Sheffield and Sheffield Hallam University, United Kingdom respectively. Her research areas include agents and data mining.

**Noor Maizura M. N.** received her Bachelor and Master degrees in Computer Science from University Putra Malaysia, Malaysia in 1994 and 1996 respectively. In 2005, she obtained PhD degree in Computer Science from The University of Manchester. Currently, she is appointed as Head Department of Computer Science, Faculty of Science and Technology, University Malaysia Terengganu. Her research interests include Decision Support Systems, Construction Tender and DNA computing.